\documentclass{article}
\usepackage{ijcai25}

\usepackage{times}
\usepackage{soul}
\usepackage{url}
\usepackage[hidelinks]{hyperref}
\usepackage[utf8]{inputenc}
\usepackage[small]{caption}
\usepackage{graphicx}
\usepackage{amsmath}
\usepackage{amsthm}
\usepackage{booktabs}
\usepackage{algorithm}
\usepackage{algorithmic}
\usepackage[switch]{lineno}
\usepackage[inline]{enumitem}
\usepackage{graphicx} %
\usepackage{amsmath,amssymb}
\usepackage{multirow}
\usepackage{array}
\usepackage{longtable}
\usepackage{tabularx}
\usepackage{multirow}

\usepackage{xspace}

\title{Exploring Equity of Climate Policies\\ using Multi-Agent Multi-Objective Reinforcement Learning}

\author{
Palok Biswas$^{*}$\and 
Zuzanna Osika$^{*}$\and 
Isidoro Tamassia$^{*}$\and 
Adit Whorra$^{*}$ \and \\
Jazmin Zatarain-Salazar\and 
Jan Kwakkel\and
Frans A. Oliehoek\And
Pradeep K. Murukannaiah\\
\affiliations
Delft University of Technology, The Netherlands\\
\{p.biswas, z.osika, j.zatarainsalazar, j.h.kwakkel, f.a.oliehoek, p.k.murukannaiah\}@tudelft.nl\\
\{isidorotamassia, aditwhorra\}@gmail.com\\
\footnotesize $^*$Equal contribution
}

\begin{document}

\maketitle

\newcommand{\justice}{\textsc{Justice}\xspace}

\begin{abstract}

Addressing climate change requires coordinated policy efforts of nations worldwide. 
These efforts are informed by scientific reports,
which rely in part on Integrated Assessment Models (IAMs), prominent tools used to assess the economic impacts of climate policies. However, traditional IAMs optimize policies based on a single objective, limiting their ability to capture the trade-offs among economic growth, temperature goals, and climate justice. As a result, policy recommendations have been criticized for perpetuating inequalities, fueling disagreements during policy negotiations. We introduce \justice, the first framework integrating IAM with Multi-Objective Multi-Agent Reinforcement Learning (MOMARL). By incorporating multiple objectives, \justice generates policy recommendations that shed light on equity while balancing climate and economic goals. Further, using multiple agents can provide a realistic representation of the interactions among the diverse policy actors. We identify equitable Pareto-optimal policies using our framework, which facilitates deliberative decision-making by presenting policymakers with the inherent trade-offs in climate and economic policy.

\end{abstract}

\section{Introduction}

Climate change poses a significant global threat, disproportionately affecting marginalized communities \cite{faus2022climate,rising2022missing}. The extent of this threat remains uncertain due to the complex interplay between climate and socioeconomic systems \cite{burke2018large,van2021optimality}. This uncertainty underscores a central challenge of climate justice: the fair allocation of burdens and benefits through the implementation of climate policies \cite{pozo2020reducing}. The complexity in assessing the implications of policy measures leads to contentious international negotiations, as seen in the Conference of the Parties (COP). Asymmetric impacts, divergent responsibilities, and differing national priorities often result in impasses and heated disagreements among policymakers \cite{wei2013review}. Climate change is thus characterized as a wicked problem in the policy domain \cite{lonngren2016systems}, necessitating interventions that seek to balance the diverse ethical preferences of stakeholders.

The Intergovernmental Panel on Climate Change (IPCC) is a key player in the global climate change discourse, and its assessment reports serve as the primary source of scientific information for international climate negotiations, such as the COP \cite{asayama2024history}. These technical reports, crafted by experts from various disciplines, including climate science and economics, provide evidence-based guidance to support effective climate action through the use of Integrated Assessment Models (IAMs) \cite{cointe2019organising}. IAMs integrate socioeconomic, climate and technological processes into a single framework to assess global mitigation pathways and simulate future socio-economic and climate scenarios.

Although IAMs are influential tools, they are criticized for suggesting inequitable mitigation policies that disadvantage developing countries \cite{rivadeneira2022justice,gambhir2022climate}. These models often simplify the complexity of climate change policies to a single objective, which is ethically problematic, as it favours dominant stakeholders and neglects non-economic metrics, such as temperature, biodiversity, and human mortality \cite{bromley1973incongruity,kasprzyk2016battling}. Recent studies seek to address these shortcomings by enhancing IAMs with multiple objectives \cite{marangoni2021adaptive,ferrari2022optimal}, or by incorporating multi-agent approaches, such as multi-regional studies \cite{zhang2022ai}. However, the multi-objective studies treat the world as a single agent, which overlooks the interests of various stakeholders. Conversely, the multi-agent approach focuses on a single objective, neglecting the diverse preferences of agents during negotiations. Currently, no single modeling framework simultaneously addresses both the multiplicity of objectives and agents, thus overlooking the complexities of the real-world. The exclusion of either multiple objectives or multiple agents limits the applicability of IAMs. Our contribution fills this gap by developing \justice, a Multi-Objective Multi-Agent Reinforcement Learning (MOMARL) IAM framework that emulates real-world negotiations and discovers equitable policy options. \footnote{Code: \href{https://github.com/pollockDeVis/JUSTICE}{https://github.com/pollockDeVis/JUSTICE}}%

MOMARL is a powerful framework for complex decision-making that requires balancing conflicting objectives and coordinating independent decisions over time. In particular, it is a crucial tool for addressing problems that involve sequential decisions \cite{radulescu2024world}. By extending reinforcement learning (RL) to accommodate multiple agents and multiple objectives, MOMARL enables trade-off management through vectorial rewards, where each component reflects performance on a specific objective. Despite the prevalence of societal problems involving multiple stakeholders with diverse preferences, the field of MOMARL remains underexplored. Simplifications such as hard-coding trade-offs or centralizing decisions limit practical applicability. Importantly, existing works are limited to theoretical examples or simple (grid-like) environments \cite{felten2024momaland,ruadulescu2020multi}.

Our contribution is twofold.
\begin{enumerate*}[label=(\arabic*)]
    \item For the IAM community, \justice introduces a multi-objective, multi-agent framework of a climate-economy model, providing a tool that can inform IPCC's synthesis reports. Such a framework can enable the design of equitable policies that account for regional disparities while balancing multiple aspects of climate change, including economic and environmental outcomes.
    \item For the MOMARL community, \justice offers a well-designed open-source implementation using the MOMALand API \cite{felten2024momaland}, allowing seamless integration with any RL algorithm. This provides a real-world testbed for evaluating and benchmarking future algorithms, supporting the development of robust and societal applications of reinforcement learning methods.
\end{enumerate*}

\section{Background}

Our interdisciplinary work is based on two foundational areas: climate modelling and reinforcement learning.

\subsection{Integrated Assessment Models}
\label{sec:iam}
IAMs offer a holistic approach to inform policy decisions on climate change \cite{gambhir2019review}. These models integrate socioeconomic, technological, and biogeochemical variables to represent interactions between simplified social and climate components \cite{rivadeneira2022justice}. IAMs primarily inform climate mitigation policies and are frequently used for policy evaluation and optimization \cite{mastrandrea2009calculating}. Originating from William Nordhaus's DICE model [\citeyear{nordhaus1992optimal}], which combined economic and climate models for global cost-benefit analyses of climate policies, IAMs analyze climate policy's impacts on both the economy and environment. DICE is a foundational framework, earning Nordhaus a Nobel Prize and being utilized by the US government to calculate the social cost of carbon, which measures the societal benefit of reducing CO\textsubscript{2} emissions \cite{grubb2021modeling}. DICE has inspired numerous IAMs and remains a valuable tool for assessing climate policies. A popular regional variation of DICE, known as RICE \cite{nordhaus1996regional}, encompasses 12 regions, and its recent and enhanced version called RICE50+ \cite{gazzotti2021persistent} expands to 57 regions for greater regional resolution.

IAMs can be broadly categorized into two types: 
\begin{enumerate*}[label=(\arabic*)]
    \item highly aggregated Cost-Benefit models and
    \item detailed process-based models \cite{van2020anticipating}.
\end{enumerate*} 
Cost-Benefit IAMs (CB-IAMs) are optimization models that identify near-term emission reduction pathways to maximize long-term benefits by considering high-level economic and climate interactions. DICE/RICE IAMs fall under this CB-optimization category. In contrast, Process-Based IAMs (PB-IAMs) are simulation models that offer detailed economic representations across multiple sectors to analyze the impacts of specific policies on economic, social, and environmental factors, with an emphasis on sector-specific environmental impacts, such as those in energy systems \cite{nikas2019detailed}. CB-IAMs are employed for global mitigation pathways, macroeconomic assessments of mitigation strategies, and strategic interactions reported by the IPCC's Working Group III (WGIII). Notably, the macroeconomic components of CB-IAMs form the foundation of many process-based models. The global mitigation pathways found by CB-IAMs can also be used to inform the simulation of PB-IAMs. The DICE/RICE family, despite its simplicity, has significantly influenced IPCC WGIII mitigation assessments in various reports, including the recent sixth assessment report, particularly in Chapters 3 and 14, which focus on Mitigation Pathways and International Cooperation, respectively \cite{IPCC2022}.

\subsection{Multi-Objective Multi-Agent RL}%

We formalize a MOMARL problem as a multi-objective multi-agent Markov decision process (MOMAMDP) with team reward \cite{ruadulescu2020multi}, defined as a tuple \(
\bigl(\mathcal{S}, \mathcal{A}, F, \mathbf{R}\bigr)
\), 
comprising \(N \geq 2\) agents and \(d \geq 2\) objectives:
\begin{itemize}
    \item \(S\) is the state space.
    \item \(\mathcal{A} = \mathcal{A}_1 \times \cdots \times \mathcal{A}_N\) is the set of joint actions, 
          where \(\mathcal{A}_n\) denotes the action set of agent \(n\).
    \item \(F: \mathcal{S} \times \mathcal{A} \to \Delta(\mathcal{S})\) is the probabilistic transition function, 
          mapping each state--joint action pair to a distribution over next states.
    \item \(\mathbf{R} = \mathbf{R}_1 \times \cdots \times \mathbf{R}_n\) represents the reward functions, 
          where \(\mathbf{R}_n: \mathcal{S} \times \mathcal{A} \times \mathcal{S} \to \mathbb{R}^d\) 
          is the vector-valued reward function for agent \(n\) across \(d\) objectives.
\end{itemize}

Agents optimize policies $\pi_n$ to maximize expected discounted returns:

\[
\mathbf{v}^\pi_n = \mathbb{E} \left[ \sum_{t=0}^\infty \gamma^t \mathbf{R}_n(s_t, \mathbf{a}_t, s_{t+1}) \mid \boldsymbol{\pi} \right]
\]
where $\boldsymbol{\pi} = ({\pi}_1, \dots, {\pi}_n)$ is the joint vector policy of the agents acting in the environment, $\gamma$ is the discount factor, and $\mathbf{R}_n(s_t, \mathbf{a}_t, s_{t+1})$ is the vectorial reward obtained by agent $n$ at timestep $t$ for the joint action $\mathbf{a}_t \in \mathcal{A}$ at state $s_t \in \mathcal{S}$.

\paragraph{Solution Set Concept} 
The value function \(\mathbf{v}^{\boldsymbol{\pi}}_n \in \mathbb{R}^d\) offers a partial ordering over policies, as it is a vector. Identifying the optimal policy requires a utility function \(u_n: \mathbb{R}^d \to \mathbb{R}\) to capture agents' preferences by mapping vectors to scalars.

The Pareto set, assuming the utility function is monotonically increasing, uses Pareto dominance (\(\succ_P\)), where a vector dominates another if it is at least equal in all objectives and strictly better in one. When agents share a team reward (\(\mathbf{v}_1^{\boldsymbol{\pi}} = \mathbf{v}_2^{\boldsymbol{\pi}} = \cdots = \mathbf{v}_n^{\boldsymbol{\pi}} = \mathbf{v}^{\boldsymbol{\pi}}\)), Pareto dominance applies directly. For a set of policies \(\boldsymbol{\Pi}\), the Pareto set \(P(\boldsymbol{\Pi})\) includes all undominated policies:
\[
P(\boldsymbol{\Pi}) = \{\boldsymbol{\pi} \in \boldsymbol{\Pi} \mid \nexists \boldsymbol{\pi}' \in \boldsymbol{\Pi} : \mathbf{v}^{\boldsymbol{\pi}'} \succ_P \mathbf{v}^{\boldsymbol{\pi}}\}.
\]

The Pareto front \(F(P)\) contains value vectors for Pareto-optimal policies in \(P(\boldsymbol{\Pi})\). When the utility function is a positively weighted linear sum, the solution set forms the convex hull of \(\mathbf{v}^{\boldsymbol{\pi}}\) \cite{felten2024multi}.

\section{The \justice Model}

The \justice IAM is a modular and efficient framework created to assess various modeling assumptions related to economic growth, damage functions, abatement costs, and social welfare. It allows experimentation with different uncertainties without incurring high computational costs. We provide an outline of the model and its MOMARL formulation.

\subsection{High-Level Outline}
\justice incorporates the economy, damage, and abatement modules from the RICE50+ model \cite{gazzotti2021persistent} and integrates them with the FAIR climate model \cite{smith2023climate}, facilitating probabilistic assessments of climate policies across various climate sensitivity scenarios. FAIR is a simplified emulator of complex climate models grounded in climate physics that accurately reproduces historical climate data and is utilized by the IPCC \cite{leach2021fairv2,IPCC2022}. Unlike other IAMs, \justice identifies Pareto-optimal policies in a multi-objective framework, assessing their robustness across various socioeconomic and climate uncertainties and considering the distributional impacts of the Pareto-optimal policies.

\justice operates on a yearly resolution (instead of a 5-year interval used in RICE50+) and covers 57 independent regions (Table 1 in Appendix A\footnote{\label{app}{Online Appendix: \href{https://doi.org/10.5281/zenodo.15424271}{https://doi.org/10.5281/zenodo.15424271}}}). \justice utilizes the Shared Socioeconomic Pathway (SSP) dataset \cite{sspriahi2017shared} for exogenous regional economic growth, carbon intensity, and population projections, and the Representative Concentration Pathway (RCP) dataset \cite{meinshausen2011rcp} informs emission trajectories for other greenhouse gases and aerosols. The SSPs and RCPs provide a consistent framework for integrating socioeconomic and climate research, representing various future scenarios through different socioeconomic narratives and climate forcings. The exogenous SSP data is denoted with a superscript $(^*)$ in the equations.

\begin{figure}[!htb]
    \centering
    \includegraphics[width=\linewidth]{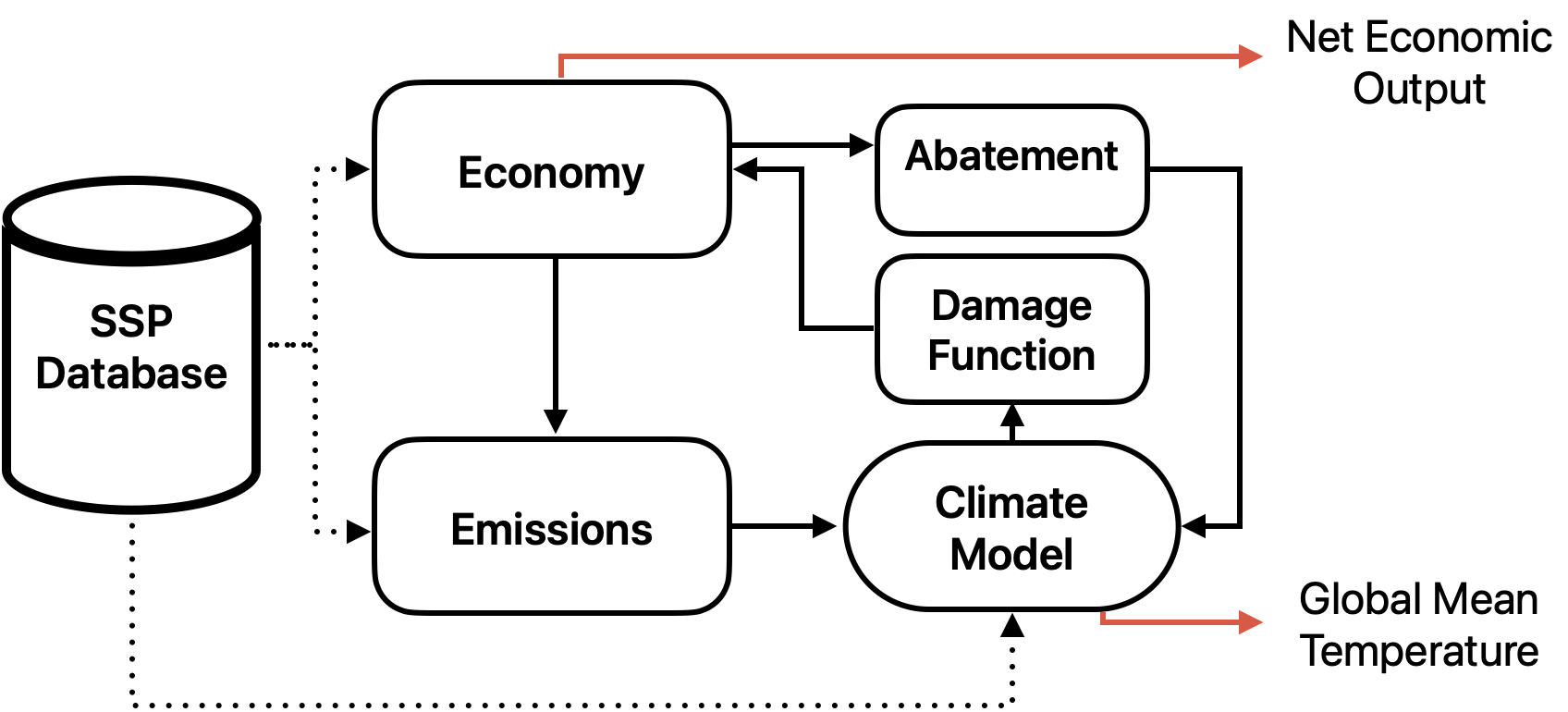}
    \caption{Overview of \justice. The main outputs of the model have been highlighted with red arrows.}
    \label{fig:justice}
\end{figure}

Figure~\ref{fig:justice} provides an overview of the \justice IAM. The economy sub-model employs the neoclassical Cobb-Douglas production function \cite{gazzotti2021persistent}, as in the DICE IAM, to compute economic growth using exogenous labour and total factor productivity data along with capital stock. The resulting economic output feeds the Emissions sub-model, which calculates CO\textsubscript{2} emissions at each timestep based on output and carbon intensity, defined as the fossil fuel share in energy production. The CO\textsubscript{2} emissions from various regions are input into the FAIR climate model, which aggregates them to calculate global mean surface temperature, expressed in °C above pre-industrial levels. This global temperature is then downscaled to regional temperature rise using a data-driven statistical downscaler. The downscaled regional temperature is used by the Damage Function to compute the fraction of economic output damaged by temperature increases in every region. Additionally, the Emissions sub-model allows for CO\textsubscript{2} mitigation, with associated costs calculated by the Abatement module, based on the emission control rate chosen by agents (or regions) at each timestep. Detailed descriptions and equations for each model component are available in Appendix B \footnotemark[2].

\subsection{Multi-Agent Multi-Objective Formulation}
We model \justice as a MOMAMDP with team reward as specified below. %

\begin{description}[leftmargin=0em,itemsep=0.5em]
    \item[Agents] 
We model 12 agents, each representing a macro-region that groups countries following RICE region specification \cite{nordhaus1996regional}. A map of these macro regions with regional mapping table is provided in Appendix A \footnotemark[2]. While \justice can generate data for 57 entities, modelling a MOMARL problem with 57 regions is computationally expensive and complicates analysis.

    \item[Actions] Each agent's action is a two-dimensional discrete vector, where the actions of agent \(n\) are: 
\begin{itemize}
    \item Emission Control Rate (ECR) \(\in \{0, 0.1, 0.2, \dots, 1.0\} \),
    \item Savings Rate (SR) \(\in \{0, 0.1, 0.2, \dots, 1.0\} \).
\end{itemize}
ECR represents the percentage of emissions removed from the atmosphere relative to the baseline SSP emissions. It is used in the Emissions module to assess emissions reduction, with the cost of this reduction being calculated in the Abatement module. SR determines the level of investment, indicating the amount of capital saved and reinvested into the economy, which in turn influences the economic output of each agent (a higher savings rate leads to higher economic output). Each agent undertakes these actions annually.

\item[Observations] The agents can observe their local outcomes,
the global temperature, and the mitigation rates
set by all the agents at the previous timestep. For \(n\)-th agent at timestep \(t\), the observation and their ranges are as follows:

\begin{equation}
    {\Omega}_{t,n} = {Y}^{GROSS}_{t,n}\cdot{\omega}_{t,n}, \quad  {\Omega}_{t,n}\in [0, \infty),
    \label{damage_costs}
\end{equation}
where \({\Omega}_{t,n}\) is economic damage by the climate change in trillion USD adjusted to 2005 Purchasing Power Parity (PPP), \( {Y}^{GROSS}_{t,n} \) is the gross economic output and \({\omega}_{t,n}\) is the damage fraction for agent \(n\) at timestep \(t\) and is calculated in the Damage module of \justice.

\begin{align}
    {\Lambda}_{t,n} &= {\zeta}_{t,n} \cdot {\varepsilon}^{*}_{t,n}  
    \cdot \left( \frac{{AC1}_{t,n}}{2} {ECR}_{t,n}^{2} + \frac{{AC2}_{t,n}}{5} {ECR}_{t,n}^{5} \right), \nonumber \\
    {\Lambda}_{t,n} &\in [0, \infty),
\label{abatement}
\end{align}

where \({\Lambda}_{t,n}\) is the abatement cost (cost for the regional economy to mitigate CO\textsubscript{2}), in trillion USD adjusted to 2005 PPP, \({\zeta}_{t,n}\) is the statistical correction factor, \({AC1}_{t,n}\)  and  \({AC2}_{t,n}\)  are the region-specific abatement coefficients of agent \(n\) at timestep \(t\) calculated in the Abatement module of \justice.

\begin{equation}
    {Y}^{NET}_{t,n} = {Y}^{GROSS}_{t,n}-{\Omega}_{t,n} - {\Lambda}_{t,n} \label{net_output}, \quad  {Y}^{NET}_{t,n}\in [0, \infty),
\end{equation}
where \( {Y}^{NET}_{t,n} \) is the net economic output  of agent \(n\) at timestep \(t\) in trillion USD adjusted to 2005 PPP and is calculated in the Economy module of \justice.

\begin{align}
    {\varepsilon}_{t,n} &= {CI^*}_{t,n} \cdot {Y}^{GROSS}_{t,n} \cdot (1 - {ECR}_{t,n}) + {AFOLU}^{*}_{t,n}, \nonumber \\ 
    {\varepsilon}_{t,n} &\in [0, \infty) \label{emissions},
\end{align}
where \({\varepsilon}_{t,n}\) represents total emissions (Annual CO\textsubscript{2} emissions in Gigatonne, GtCO\textsubscript{2} per year), \({CI^*}_{t,n}\) is the carbon intensity (exogenous SSP data), \({AFOLU}^{*}_{t,n}\) is the Agriculture, Forestry, and Other Land Uses emissions (exogenous SSP data) of agent \(n\) at timestep \(t\) and is calculated in the Emissions module of \justice.
\begin{equation}
    {GMT}_{t} = \textit{FAIR}\left(\sum_{n\in N} {\varepsilon}_{{t-1},n} \right), \quad  {GMT}_{t}\in [0, \infty),  
    \label{fair}
\end{equation}
where \( {GMT}_{t}\) is the rise in the global average surface temperature since the pre-industrial era at timestep \(t\) (in degrees Celsius) outputted by the FAIR model in the climate module.

\begin{equation}
\begin{split}
    {RMT}_{t,n} &= {TC1}_{t,n} + {TC2}_{t,n} \cdot {GMT}_{t}, \\ 
    {RMT}_{t,n} &\in [0, \infty),
\end{split}
\end{equation}

where \( {RMT}_{t,n}\) is the rise in the regional average surface temperature since the pre-industrial era (in degrees Celsius), \({TC1}_{t,n}\)  and  \({TC2}_{t,n}\) are the region-specific downscaler coefficients for agent \(n\) at timestep \(t\) and is calculated in the Downscaler of the Climate module.

\begin{equation}
\begin{split}
    \mathbf{{VECR}_{t,n}} &= \{\text{ECR}_{t-1,1}, \text{ECR}_{t-1,2}, \dots, \text{ECR}_{t-1,N} \}, \\
    \mathbf{{VECR}_{t,n}} &\in \{0, 0.1, 0.2, \dots, 1.0\}^{N},
\end{split}
\end{equation}
where $\mathbf{{VECR}_{t,n}}$ is a vector of emission control rates adopted by the agents at the previous timestep that agent \(n\) observes at timestep \(t\).

    \item[Rewards] The rewards are modelled as team rewards, which assume collaboration among agents, where all agents receive the same return vector for executing the policy, reflecting their shared goal of improving global outcomes. This setup can be easily adjusted to individual rewards if needed. In our approach, each agent receives a two-dimensional vector of continuous rewards based on the following metrics:

\begin{itemize}
    \item \textbf{Inverse Global Temperature (IGT):} 
    \begin{equation}
    {IGT}_{t,n} = 1/ {GMT}_{t}, \quad  {IGT}_{t,n}\in [0, 1].  
    \label{igt}
\end{equation}

    \item \textbf{Global Economic Output (GEO):} 
    \begin{equation}
    {GEO}_{t,n} = \sum_{n=1}^{N} {Y}^{NET}_{t,n}, \quad  {GEO}_{t,n}\in [0, \infty).
    \label{geo}
\end{equation}    
\end{itemize}

In the RL setup, agents maximize their reward. Thus, we use the inverse of the Global Mean Temperature to reflect the goal of minimizing temperature rise. Similarly, we use the absolute value of the net economic output to reflect our aim of maximizing economic performance.

    \item[Starting State] The \justice simulation is initialized in 2015, using the SSP-2 data. SSP-2 is commonly used in IAM literature and it is the continuation of current emission trends into the future.
    
    \item[Episode Termination] Each episode consists of 285 timesteps, with each timestep representing one year, starting from 2015 and continuing until 2300. This extended timeframe accounts for the lag effects of the climate response, allowing the evaluation of damages not only for the current period but also for the centuries that follow.

\end{description}

\section{Experimental Setup}

We train our agents using Multi-Objective Multi-gent Proximal Policy Optimization (MOMAPPO)\cite{felten2024momaland}, an extension of Multi-Agent PPO (MAPPO) \cite{yu2022surprising} designed for multi-objective settings. MOMAPPO decomposes a multi-objective problem into multiple single-objective problems using a weighted-sum scalarization function for simplicity. Rewards are normalized to ensure consistency across objectives, and 100 weight vectors are uniformly sampled to generate diverse solutions. For each weight, MOMAPPO trains a multi-agent policy using MAPPO, evaluating its performance and adding non-dominated policies to the solution set. MOMAPPO is trained for 1 million global steps, with evaluations every (approximately) 20,000 steps for 10 random seeds. Appendix C\footnotemark[2] includes additional details. %

\subsection{Performance Indicators}

Evaluating and comparing solution sets in MOMARL is more complex than in single-agent, single-objective RL due to the lack of inherent ordering of solution sets and intertwined performance across objectives. This added complexity leads to varied evaluation methods in MOMARL. To study convergence, we use two most commonly used approaches from both MORL and MARL domains for their suitability in MOMARL settings \cite{hayes2022practical} and to study the equality of distribution between agents we use GINI index:

\begin{description}[leftmargin=0em,itemsep=0.1em]
    \item[\textit{Hypervolume} ($\uparrow$)] \cite{zitzler2003performance} represents the region or (hyper-)volume between the points in the solution set and a reference point. The reference point indicates the lower bound for each objective. 
    A solution set can be assessed by comparing its hypervolume with that of competing algorithms or the true Pareto front, if known.
    
    \item[\textit{Expected utility} ($\uparrow$)] When the decision-maker's utility function \(u\) is linear, the expected utility (EU) metric \cite{ruadulescu2020multi} can be used to represent the expected utility over a distribution of reward weights \(W\).

    \item[\textit{GINI Index} ($\downarrow$)] The GINI index is a measure of inequality that quantifies disparities among agents based on a specific metric, with 0 indicating perfect equality and 1 indicating maximal inequality; we employ Concept-1 GINI by \cite{milanovic2011worlds} to assess international inequality, where each agent represents a region and the index reflects whether their metrics are converging. 
\end{description}

\subsection{RICE50+ Comparison}
We compare our results with the default RICE50+ model, as outlined in Section~\ref{sec:iam}. This model consists of 57 regions and integrates optimization directly within the global climate simulation. The optimization follows a single-objective approach, employing a social welfare function as the global objective and using a single representative agent to optimize welfare, rather than a multi-agent multi-objective framework. The results presented in the next section are obtained using standard IAM methods, specifically non-linear programming within the General Algebraic Modeling Language (GAMS) for policy optimization.  Unlike our approach, RICE50+ does not perform inter-temporal optimization at each timestep. Instead, it optimizes over the entire time horizon by using a social welfare function as a proxy for consumption per capita. Since consumption per capita is directly derived from net economic output (Equation 16a in Appendix B\footnotemark[2]), we extract the net economic output from RICE50+ to compare with our objectives.

\section{Results}
The objectives of our experiments are to 
\begin{enumerate*}[label=(\arabic*)]
    \item verify the convergence of \justice MOMARL, 
    \item compare the solutions between \justice and the RICE50+ model, and
    \item analyze key solutions from the Pareto set.
\end{enumerate*}
We present results in line with these objectives.

\subsection{Convergence}

\begin{figure}[!htb]
    \centering
    \includegraphics[width=\columnwidth]{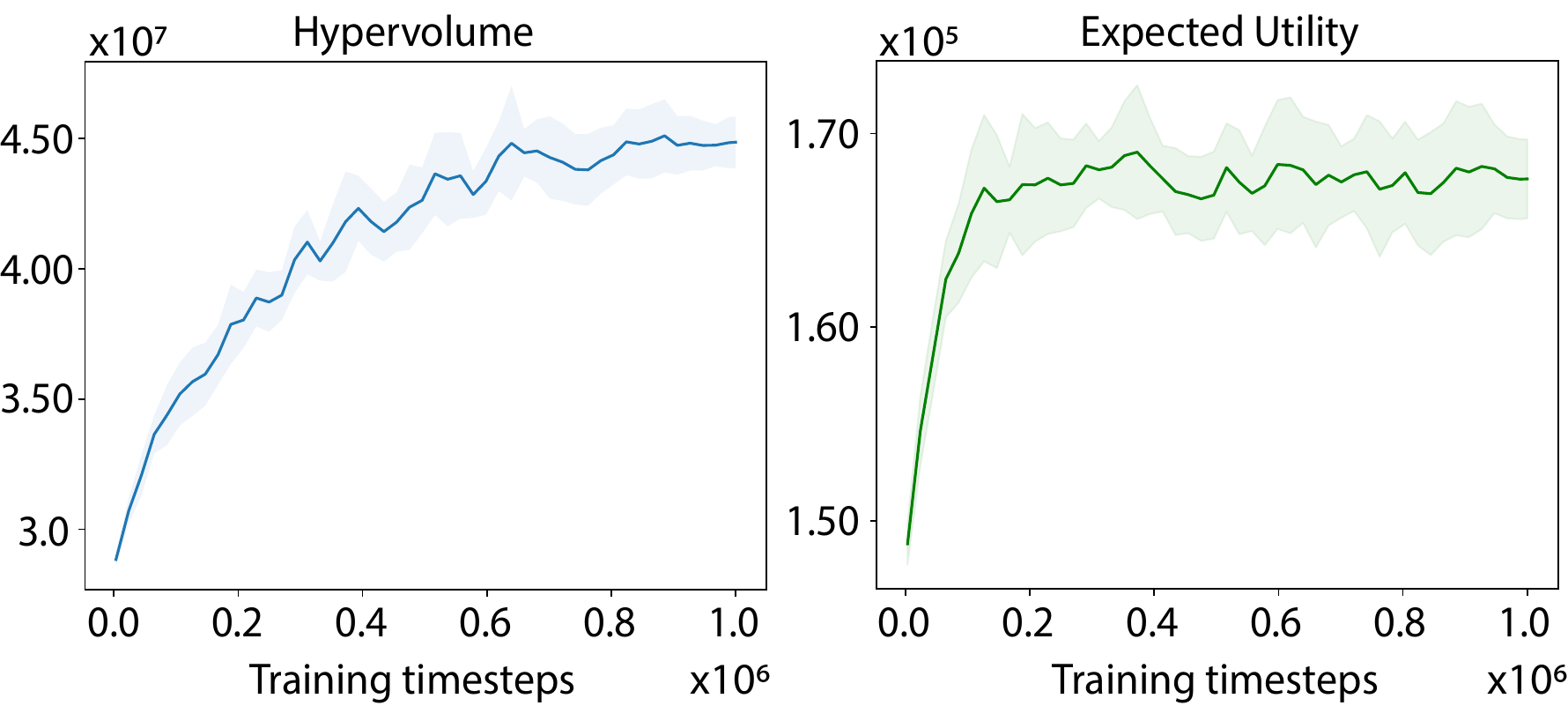} 
    \caption{Mean hypervolume and expected utility over training steps (shaded area represents standard deviation).}
    \label{fig:perf}
\end{figure}

Figure~\ref{fig:perf} shows that our agents converge and demonstrate consistent training over time, with both performance metrics growing and eventually stabilizing. Note that both the hypervolume and Expected Utility metrics appear large due to their exponential scaling with the number of objectives and the achievable value ranges, particularly influenced by the high values of the GEO objective.

\subsection{Solutions}

\begin{figure}[!htb]
    \centering
    \includegraphics[width=0.9\columnwidth]{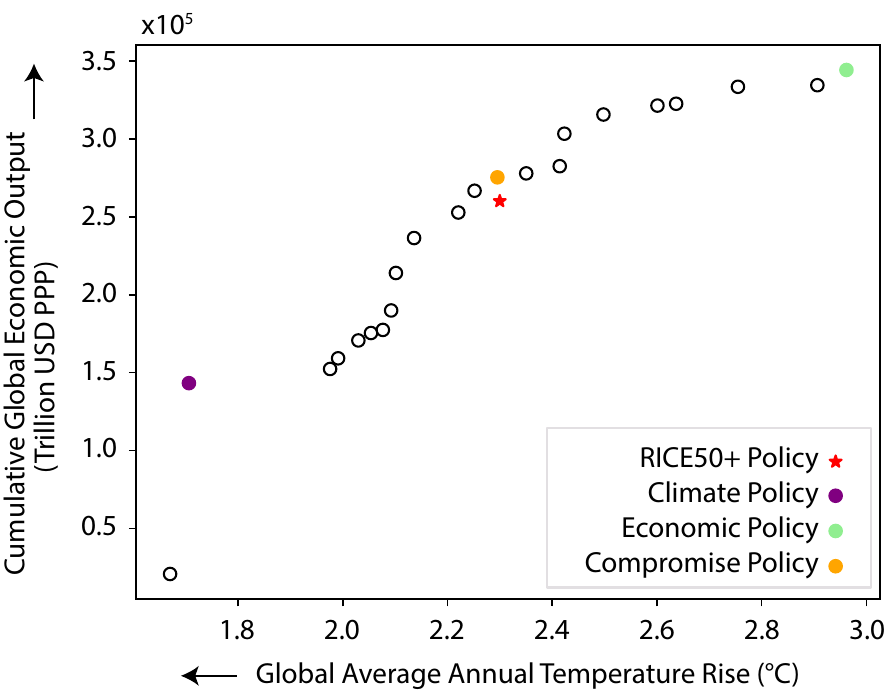} 
    \caption{Pareto set of policies obtained by \justice (across 10 random seeds) with the RICE50+ polict for comparison. Arrows indicate the direction of preference for the objectives.}%
    \label{fig:solution-set}
\end{figure}

Figure~\ref{fig:solution-set} shows the Pareto set of solutions produced by \justice and the single RICE50+ solution (red star). We transform the objective values for simplicity: Total Global Economic Output represents the GEO objective, and Global Average Annual Temperature corresponds to the mean of the inverse of IGT (inverted to retrieve the temperature). 

\justice produces 22 solutions (policies), shown as circles in Figure~\ref{fig:solution-set}. Among these, three are highlighted: Climate Policy (purple), Economic Policy (green), and Compromise Policy (yellow). Although the purple policy is not the most extreme in terms of climate performance, it achieves substantial economic gains with only a slight increase in temperature compared to the extreme climate solution. Therefore, it is chosen as the Climate Policy. This example illustrates how the multi-objective approach supports decision-making by presenting a range of trade-offs.

The RICE50+ solution lies close to the \justice Pareto front, roughly in the middle, indicating a balance between economy and  temperature rise. However, the \justice Compromise Policy dominates the RICE50+ solution, albeit by a small margin. Thus,  \justice not only offers a range of solutions but also a slightly better solution than RICE50+.

\subsection{Comparison of Key Solutions}
\begin{figure*}[t] %
    \centering
    \includegraphics[width=0.85\textwidth]{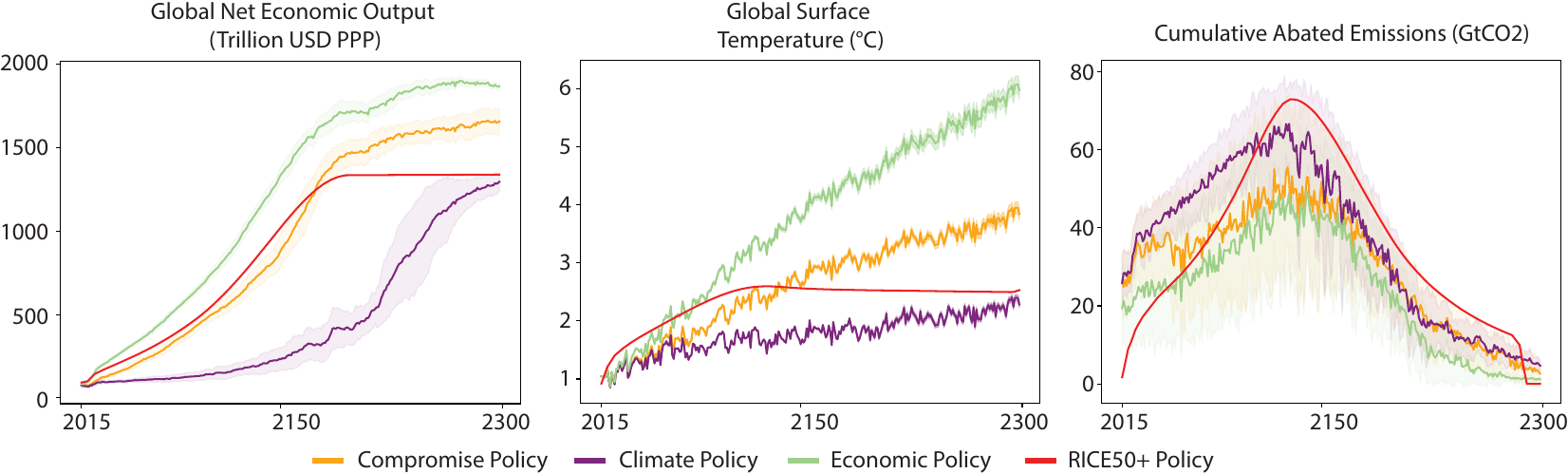} %
    \caption{Performance of selected \justice policies over time (years on x-axis) for three important indicators: Global Net Economic Output, Temperature, and Cumulative Abated Emissions. The indicators show mean values and standard deviation (over 10 seeds) for selected \justice and RICE50+ policies over time. }
    \label{fig:results-details}
\end{figure*}

We perform a comprehensive analysis of the highlighted policies in Figure~\ref{fig:solution-set}---the three \justice policies and the RICE50+ policy. To do so, we plot the key IAM outcomes from 2015 to 2300 for these policies in Figure~\ref{fig:results-details}. 

\begin{description}[leftmargin=0em,nosep]
\item [The economic output] trajectories are shown in left panel of Figure~\ref{fig:results-details}. As expected, the best economic policy achieves the highest output. The best climate policy starts with the lowest output but eventually aligns with RICE50+ levels, as its rapid and deep mitigation efforts stabilize the climate and reduce damages, stimulating economic growth after 2150. The compromise policy presents an interesting trade-off; although it sacrifices slightly more output in the near term compared to RICE50+, it leads to significantly higher long-term growth, surpassing RICE50+ by hundreds of trillions of dollars, by reducing climate damages through immediate mitigation.

\item [The temperature] trajectories are shown in the center panel of Figure~\ref{fig:results-details}. As expected, the best economic policy results in the highest temperature rise, while the best climate policy yields the lowest. Under the compromise policy, temperatures exceed 2°C by the end of the 2100s, reaching about 3.5°C by 2300. %
The RICE50+ temperature projections are significantly lower than those of the \justice compromise policy over the long term. This discrepancy highlights a common criticism of cost-benefit IAMs: their use of simple climate models inadequately captures the complexities and uncertainties of climate sensitivity, leading to smooth projections that miss abrupt, high-impact events \cite{mastrandrea2001integrated,stanton2009inside} temperature rises, leading to policy inertia and delayed action \cite{fussellogical,stern2022time}.  We address these issues by using the FAIR model, which captures nonlinear dynamics and feedback mechanisms, effectively handling uncertainty in climate sensitivity (see Appendix B\footnotemark[2]). FAIR produces temperature scenarios that align with more complex Earth System Models featured in the IPCC \cite{cmipo2016scenario}.

\item [The abated emissions] trajectories are shown in the right panel of Figure~\ref{fig:results-details}.
All three \justice policies favour rapid near-term mitigation, unlike RICE50+, which delays mitigation and peaks only around 2150. This finding aligns with the criticisms of traditional CB-IAMs for undervaluing early mitigation efforts. Among \justice policies, the best climate policy (purple trajectory) achieves the highest cumulative abated emissions. Both the best economic and compromise policies show similar mitigation levels, though the compromise policy emphasizes earlier action. This finding also aligns with the IPCC's recommendations for rapid near-term mitigation to limit global warming and promote sustainable growth with minimal climate damage \cite{IPCC2022}.
\end{description}

\subsection{Equity Analysis}
\begin{figure}[!htb]
    \centering
    \includegraphics[width=0.65\columnwidth]{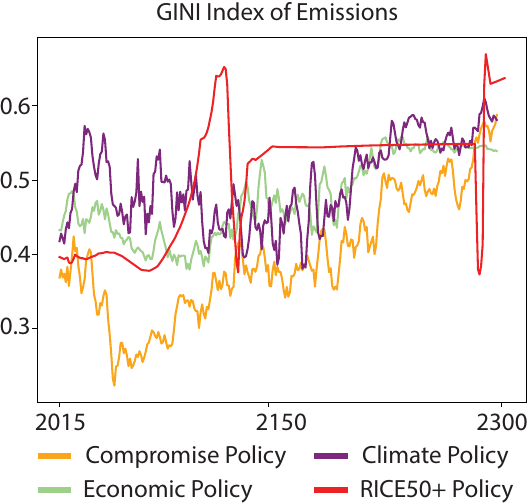} 
    \caption{GINI Index of Emissions Over Time for 12 Regions (years on x-axis).}
    \label{fig:gini}
\end{figure}

Figure~\ref{fig:gini} shows the GINI index for emissions among 12 macro-regional agents, illustrating the distribution of mitigation and the future emissions budget. The emissions metric is used because it captures the agent's primary action of emissions control. A basic setup in JUSTICE is employed without explicit equity objectives (such as Utilitarianism in RICE50+) to examine how disaggregating objectives and including multiple agents affect equity outcomes. For consistency, RICE50+ results from 57 regions have been aggregated to match our 12-region model.

As seen in Figure~\ref{fig:gini}, the best climate and economic policies yield higher emissions inequality than the compromise policy. In the climate policy, inequality tends to be higher on average compared to other policies, as developing countries are potentially required to limit emissions despite their growth needs. In contrast, the GINI index for the economic policy is higher than the compromise because high-emitting developed countries are likely to increase emissions to drive economic growth. The compromise policy has the lowest average GINI index and a more equitable emissions distribution compared to RICE50+, which, despite a similar starting point, sees inequality rise sharply to around 0.5 by 2100. This setup demonstrates that the objectives that agents optimize and the chosen Pareto-optimal policy shape the allocation of mitigation burdens and emissions budgets across agents, setting the MOMARL approach apart from single-objective models. Fair policy design requires distinct equity objectives that convert outputs from \justice, such as regional damage and abatement costs, into an equity score while further disaggregating regions to better represent smaller nations. Nonetheless, the MOMARL approach is a promising step towards equitable policy design.

\section{Discussion}

Our results demonstrate how the optimization setup in IAMs can significantly affect the resulting climate policy recommendations. We compare \justice with the traditional single-objective optimization of RICE50+, a successor of widely used RICE IAM \cite{nordhaus1996regional} for studying optimal mitigation pathways. Our findings underscore the advantages of multi-objective optimization, which offers a range of solutions compared to single-objective optimization. We do not make any policy recommendations in this study; rather, we illustrate how policy outcomes are sensitive to the framing of the optimization problem and highlight how equity considerations vary with different solutions and objectives. This flexibility is crucial when agents have unequal economic capacities and face disproportionate climate impacts in an uncertain future.

\justice can have real-world impact in several ways.

\begin{description}[leftmargin=0em,itemsep=0.1em]
\item [Decision Support] The multi-objective multi-agent nature makes \justice a powerful decision-support tool for climate (e.g., COP) negotiations. In this use case, each \justice agent can represent a country or a coalition with associated preferences, simulating various policy options and implications. This enables stakeholders to propose, reassess, and refine actions, promoting transparent and robust negotiations.

\item [Scientific Discourse] Our model can enrich scientific (e.g., IPCC) reports by providing complementary insights that capture dimensions often overlooked by conventional models. By exploring additional strategies through multi-objective multi-agent reinforcement learning, our approach broadens the scope of potential solutions and offers innovative pathways for regions to understand the consequences of their actions and implement more equitable climate policies.

\item [Broader Engagement] While IAMs are crucial for policy recommendations, their technical nature and costly software licenses (e.g., models written in GAMS) often limit accessibility. By offering our framework as open source, we empower stakeholders to explore different scenarios and policy impacts. A more user-friendly interface could enhance this further. Further, our approach bridges research (e.g., IAM and AI) communities, demonstrating the benefits of building on models and algorithms developed by each other.%
\end{description}

\subsection{Limitations and Directions}

The following are key limitations of our work. Addressing these limitations opens interesting avenues for future work.

\begin{description}[leftmargin=0em,itemsep=0.1em]
\item [Reward and Utility] The MOMARL algorithm we employ makes two key simplifying assumptions. First, a team-based reward structure, where all agents receive the same global reward. This is a simplification since, in practice, countries are likely to prioritize their individual interests. Second, it requires a linear utility function for the multi-objective aspect, which simplifies real-world scenarios. To our knowledge, this is the only MOMARL algorithm suitable for our setup of continuous state and reward spaces, and a discrete action space (see Table 2 in \cite{felten2024momaland}). Relaxing these assumptions, i.e., exploring individual reward structures and other utility functions, is an important direction for future work.

\item [Scalability] We perform analyses considering 12 macro-regions of the world. Although this is a substantial improvement over the single-agent assumption of traditional IAMs, a more fine-grained resolution may be necessary to derive insights on equity. This requires a substantial speedup of current MOMARL algorithms. One direction in this regard is to experiment with alternative weight-sampling techniques, such as adaptive weight sampling \cite{felten2024multi}. %

\item [Communication] Our current model assumes cooperation among regions, whereas real climate negotiations involve complex bargaining and conflicting agendas. Exploring different communication strategies among agents, and modeling trade and capital transfers, could improve the model’s ability to reflect real-world dynamics. Further, training the agents on different climate ensembles of FAIR, which represent varying plausible climate sensitivities, would allow the agents to take actions that are robust under climate uncertainty. 

\item [Policy Implementation] We acknowledge that the policies generated by \justice may be unrealistic, as both RICE50+ and \justice assume mitigation begins at the start of the model run, and the drastic mitigation suggested by \justice is infeasible due to geopolitical dynamics, technological inertia, and policy implementation challenges. However, the utility of \justice is in providing a range of alternatives, which can serve as a scientific basis for stakeholder deliberations.

\end{description}

\subsection{Conclusions}

We introduce the \justice framework, the first to integrate IAM with MOMARL. Our approach stands apart from the models typically featured in climate reports---models upon which policymakers rely to inform decisions and negotiate actions. Unlike traditional models, which often oversimplify climate change into problems with a single agent and objective, our model embraces the complexity of the real world by acknowledging its multi-agent, multi-objective nature. Our experiments show that \justice produces flexible policies and allows detailed exploration of equity compared to RICE50+. Additionally, our model is open-source, implemented in Python, and designed to be accessible for exploration and experimentation of RL algorithms for an important real-world problem.

\section*{Contribution Statement}
Palok Biswas, Zuzanna Osika, Isidoro Tamassia, and Adit Whorra made equal contributions to this study and are designated as co-first authors. Jazmin Zatarain Salazar, Jan Kwakkel, Frans A. Oliehoek and Pradeep K. Murukannaiah, reviewed the final manuscript.

\bibliographystyle{named}
\bibliography{literature}

@article{felten2024momaland,
  title={MOMAland: A Set of Benchmarks for Multi-Objective Multi-Agent Reinforcement Learning},
  author={Felten, Florian and Ucak, Umut and Azmani, Hicham and Peng, Gao and R{\"o}pke, Willem and Baier, Hendrik and Mannion, Patrick and Roijers, Diederik M and Terry, Jordan K and Talbi, El-Ghazali and others},
  journal={arXiv preprint arXiv:2407.16312},
  year={2024}
}

@article{yu2022surprising,
  title={The surprising effectiveness of ppo in cooperative multi-agent games},
  author={Yu, Chao and Velu, Akash and Vinitsky, Eugene and Gao, Jiaxuan and Wang, Yu and Bayen, Alexandre and Wu, Yi},
  journal={Advances in Neural Information Processing Systems},
  volume={35},
  pages={24611--24624},
  year={2022}
}

@inproceedings{radulescu2024world,
  title={The World is a Multi-Objective Multi-Agent System: Now What?},
  author={Radulescu, Roxana},
  booktitle={27th European Conference on Artificial Intelligence},
  pages={32--38},
  year={2024},
  organization={IOS Press}
}

@article{hayes2022practical,
  title={A practical guide to multi-objective reinforcement learning and planning},
  author={Hayes, Conor F and R{\u{a}}dulescu, Roxana and Bargiacchi, Eugenio and K{\"a}llstr{\"o}m, Johan and Macfarlane, Matthew and Reymond, Mathieu and Verstraeten, Timothy and Zintgraf, Luisa M and Dazeley, Richard and Heintz, Fredrik and others},
  journal={Autonomous Agents and Multi-Agent Systems},
  volume={36},
  number={1},
  pages={26},
  year={2022},
  publisher={Springer}
}

@article{felten2024multi,
  title={Multi-Objective Reinforcement Learning based on Decomposition: A taxonomy and framework},
  author={Felten, Florian and Talbi, El-Ghazali and Danoy, Gr{\'e}goire},
  journal={Journal of Artificial Intelligence Research},
  volume={79},
  pages={679--723},
  year={2024}
}

@article{ruadulescu2020multi,
  title={Multi-objective multi-agent decision making: a utility-based analysis and survey},
  author={R{\u{a}}dulescu, Roxana and Mannion, Patrick and Roijers, Diederik M and Now{\'e}, Ann},
  journal={Autonomous Agents and Multi-Agent Systems},
  volume={34},
  number={1},
  pages={10},
  year={2020},
  publisher={Springer}
}

@article{zitzler2003performance,
  title={Performance assessment of multiobjective optimizers: An analysis and review},
  author={Zitzler, Eckart and Thiele, Lothar and Laumanns, Marco and Fonseca, Carlos M and Da Fonseca, Viviane Grunert},
  journal={IEEE Transactions on evolutionary computation},
  volume={7},
  number={2},
  pages={117--132},
  year={2003},
  publisher={IEEE}
}

@article{lonngren2016systems,
  title={Systems thinking for dealing with wicked sustainability problems: Beyond functionalist approaches},
  author={L{\"o}nngren, Johanna and Svanstr{\"o}m, Magdalena},
  journal={New developments in engineering education for sustainable development},
  pages={151--160},
  year={2016},
  publisher={Springer}
}

@article{rivadeneira2022justice,
  title={(In) justice in modelled climate futures: A review of integrated assessment modelling critiques through a justice lens},
  author={Rivadeneira, Natalia Rubiano and Carton, Wim},
  journal={Energy Research \& Social Science},
  volume={92},
  pages={102781},
  year={2022},
  publisher={Elsevier}
}

@article{ferrari2022optimal,
  title={From optimal to robust climate strategies: expanding integrated assessment model ensembles to manage economic, social, and environmental objectives},
  author={Ferrari, Luca and Carlino, Angelo and Gazzotti, Paolo and Tavoni, Massimo and Castelletti, Andrea},
  journal={Environmental Research Letters},
  volume={17},
  number={8},
  pages={084029},
  year={2022},
  publisher={IOP Publishing}
}

@article{burke2018large,
  title={Large potential reduction in economic damages under UN mitigation targets},
  author={Burke, Marshall and Davis, W Matthew and Diffenbaugh, Noah S},
  journal={Nature},
  volume={557},
  number={7706},
  pages={549--553},
  year={2018},
  publisher={Nature Publishing Group UK London}
}

@article{van2021optimality,
  title={On the optimality of 2° C targets and a decomposition of uncertainty},
  author={van der Wijst, Kaj-Ivar and Hof, Andries F and van Vuuren, Detlef P},
  journal={Nature communications},
  volume={12},
  number={1},
  pages={2575},
  year={2021},
  publisher={Nature Publishing Group UK London}
}

@article{rising2022missing,
  title={The missing risks of climate change},
  author={Rising, James and Tedesco, Marco and Piontek, Franziska and Stainforth, David A},
  journal={Nature},
  volume={610},
  number={7933},
  pages={643--651},
  year={2022},
  publisher={Nature Publishing Group UK London}
}

@article{faus2022climate,
  title={The climate change--inequality nexus: towards environmental and socio-ecological inequalities with a focus on human capabilities},
  author={Faus Onbargi, Alexia},
  journal={Journal of Integrative Environmental Sciences},
  volume={19},
  number={1},
  pages={163--170},
  year={2022},
  publisher={Taylor \& Francis}
}

@article{wei2013review,
  title={Review of proposals for an Agreement on Future Climate Policy: Perspectives from the responsibilities for GHG reduction},
  author={Wei, Yi-Ming and Zou, Le-Le and Wang, Kai and Yi, Wen-Jin and Wang, Lu},
  journal={Energy Strategy Reviews},
  volume={2},
  number={2},
  pages={161--168},
  year={2013},
  publisher={Elsevier}
}

@article{stanton2009inside,
  title={Inside the integrated assessment models: Four issues in climate economics},
  author={Stanton, Elizabeth A and Ackerman, Frank and Kartha, Sivan},
  journal={Climate and Development},
  volume={1},
  number={2},
  pages={166--184},
  year={2009},
  publisher={Taylor \& Francis}
}

@article{marangoni2021adaptive,
  title={Adaptive mitigation strategies hedge against extreme climate futures},
  author={Marangoni, Giacomo and Lamontagne, Jonathan R and Quinn, Julianne D and Reed, Patrick M and Keller, Klaus},
  journal={Climatic Change},
  volume={166},
  number={3},
  pages={37},
  year={2021},
  publisher={Springer}
}

@article{leach2021fairv2,
  title={FaIRv2. 0.0: a generalized impulse response model for climate uncertainty and future scenario exploration},
  author={Leach, Nicholas J and Jenkins, Stuart and Nicholls, Zebedee and Smith, Christopher J and Lynch, John and Cain, Michelle and Walsh, Tristram and Wu, Bill and Tsutsui, Junichi and Allen, Myles R},
  journal={Geoscientific Model Development},
  volume={14},
  number={5},
  pages={3007--3036},
  year={2021},
  publisher={Copernicus Publications G{\"o}ttingen, Germany}
}

@article{gazzotti2021persistent,
  title={Persistent inequality in economically optimal climate policies},
  author={Gazzotti, Paolo and Emmerling, Johannes and Marangoni, Giacomo and Castelletti, Andrea and Wijst, Kaj-Ivar van der and Hof, Andries and Tavoni, Massimo},
  journal={Nature Communications},
  volume={12},
  number={1},
  pages={3421},
  year={2021},
  publisher={Nature Publishing Group UK London}
}

@article{smith2023climate,
  title={Climate uncertainty impacts on optimal mitigation pathways and social cost of carbon},
  author={Smith, Christopher J and Al Khourdajie, Alaa and Yang, Pu and Folini, Doris},
  journal={Environmental Research Letters},
  volume={18},
  number={9},
  pages={094024},
  year={2023},
  publisher={IOP Publishing}
}

@book{milanovic2011worlds,
  title={Worlds apart: Measuring international and global inequality},
  author={Milanovic, Branko},
  year={2011},
  publisher={Princeton University Press},
  chapter={3},
  pages={20--27}
}

@article{bromley1973incongruity,
  title={On the incongruity of program objectives and project evaluation: An example from the reclamation program},
  author={Bromley, Daniel W and Beattie, Bruce R},
  journal={American Journal of Agricultural Economics},
  volume={55},
  number={3},
  pages={472--476},
  year={1973},
  publisher={Wiley Online Library}
}

@article{nordhaus1992optimal,
  title={An optimal transition path for controlling greenhouse gases},
  author={Nordhaus, William D},
  journal={Science},
  volume={258},
  number={5086},
  pages={1315--1319},
  year={1992},
  publisher={American Association for the Advancement of Science}
}

@article{grubb2021modeling,
  title={Modeling myths: On DICE and dynamic realism in integrated assessment models of climate change mitigation},
  author={Grubb, Michael and Wieners, Claudia and Yang, Pu},
  journal={Wiley Interdisciplinary Reviews: Climate Change},
  volume={12},
  number={3},
  pages={e698},
  year={2021},
  publisher={Wiley Online Library}
}

@article{van2020anticipating,
  title={Anticipating futures through models: the rise of Integrated Assessment Modelling in the climate science-policy interface since 1970},
  author={Van Beek, Lisette and Hajer, Maarten and Pelzer, Peter and van Vuuren, Detlef and Cassen, Christophe},
  journal={Global Environmental Change},
  volume={65},
  pages={102191},
  year={2020},
  publisher={Elsevier}
}

@article{cointe2019organising,
  title={Organising policy-relevant knowledge for climate action: integrated assessment modelling, the IPCC, and the emergence of a collective expertise on socioeconomic emission scenarios},
  author={Cointe, B{\'e}atrice and Cassen, Christophe and Nada{\"\i}, Alain},
  journal={Science \& Technology Studies},
  year={2019},
  publisher={European Association for the Study of Science and}
}

@article{asayama2024history,
  title={The history and future of IPCC special reports: A dual role of politicisation and normalisation},
  author={Asayama, Shinichiro},
  journal={Climatic Change},
  volume={177},
  number={9},
  pages={137},
  year={2024},
  publisher={Springer}
}

@article{gambhir2019review,
  title={A review of criticisms of integrated assessment models and proposed approaches to address these, through the lens of BECCS},
  author={Gambhir, Ajay and Butnar, Isabela and Li, Pei-Hao and Smith, Pete and Strachan, Neil},
  journal={Energies},
  volume={12},
  number={9},
  pages={1747},
  year={2019},
  publisher={MDPI}
}

@article{mastrandrea2009calculating,
  title={Calculating the benefits of climate policy: examining the assumptions of integrated assessment models},
  author={Mastrandrea, Michael D},
  journal={Pew Center on Global Climate Change Working Paper},
  year={2009},
  publisher={Citeseer}
}

@article{gambhir2022climate,
  title={Climate change mitigation scenario databases should incorporate more non-IAM pathways},
  author={Gambhir, Ajay and Ganguly, Gaurav and Mittal, Shivika},
  journal={Joule},
  volume={6},
  number={12},
  pages={2663--2667},
  year={2022},
  publisher={Elsevier}
}

@article{kasprzyk2016battling,
  title={Battling arrow’s paradox to discover robust water management alternatives},
  author={Kasprzyk, Joseph R and Reed, Patrick M and Hadka, David M},
  journal={Journal of Water Resources Planning and Management},
  volume={142},
  number={2},
  pages={04015053},
  year={2016},
  publisher={American Society of Civil Engineers}
}

@article{zhang2022ai,
  title={AI for global climate cooperation: modeling global climate negotiations, agreements, and long-term cooperation in RICE-N},
  author={Zhang, Tianyu and Williams, Andrew and Phade, Soham and Srinivasa, Sunil and Zhang, Yang and Gupta, Prateek and Bengio, Yoshua and Zheng, Stephan},
  journal={arXiv preprint arXiv:2208.07004},
  year={2022}
}

@article{nikas2019detailed,
  title={A detailed overview and consistent classification of climate-economy models},
  author={Nikas, Alexandros and Doukas, Haris and Papandreou, Andreas},
  journal={Understanding risks and uncertainties in energy and climate policy: Multidisciplinary methods and tools for a low carbon society},
  pages={1--54},
  year={2019},
  publisher={Springer International Publishing}
}

@misc{IPCC2022,
  title     = {Climate Change 2022: Mitigation of Climate Change. Contribution of Working Group III to the Sixth Assessment Report of the Intergovernmental Panel on Climate Change},
  author    = {P.R. Shukla and J. Skea and R. Slade and A. Al Khourdajie and R. van Diemen and D. McCollum and M. Pathak and S. Some and P. Vyas and R. Fradera and M. Belkacemi and A. Hasija and G. Lisboa and S. Luz and J. Malley},
  year      = {2022},
  institution = {Intergovernmental Panel on Climate Change},
  publisher = {Cambridge University Press},
  address   = {Cambridge, UK and New York, NY, USA},
  doi       = {10.1017/9781009157926},

}

@article{nordhaus1996regional,
  title={A regional dynamic general-equilibrium model of alternative climate-change strategies},
  author={Nordhaus, William D and Yang, Zili},
  journal={The American Economic Review},
  pages={741--765},
  year={1996},
  publisher={JSTOR}
}

@article{pozo2020reducing,
  title={Reducing global environmental inequality: Determining regional quotas for environmental burdens through systems optimisation},
  author={Pozo, Carlos and Gal{\'a}n-Mart{\'\i}n, A and Cort{\'e}s-Borda, D and Sales-Pardo, Marta and Azapagic, Adisa and Guimer{\`a}, R and Guill{\'e}n-Gos{\'a}lbez, Gonzalo},
  journal={Journal of cleaner production},
  volume={270},
  pages={121828},
  year={2020},
  publisher={Elsevier}
}

@article{sspriahi2017shared,
  title={The Shared Socioeconomic Pathways and their energy, land use, and greenhouse gas emissions implications: An overview},
  author={Riahi, Keywan and Van Vuuren, Detlef P and Kriegler, Elmar and Edmonds, Jae and O’neill, Brian C and Fujimori, Shinichiro and Bauer, Nico and Calvin, Katherine and Dellink, Rob and Fricko, Oliver and others},
  journal={Global environmental change},
  volume={42},
  pages={153--168},
  year={2017},
  publisher={Elsevier}
}

@article{meinshausen2011rcp,
  title={The RCP greenhouse gas concentrations and their extensions from 1765 to 2300},
  author={Meinshausen, Malte and Smith, Steven J and Calvin, Katherine and Daniel, John S and Kainuma, Mikiko LT and Lamarque, Jean-Francois and Matsumoto, Kazuhiko and Montzka, Stephen A and Raper, Sarah CB and Riahi, Keywan and others},
  journal={Climatic change},
  volume={109},
  pages={213--241},
  year={2011},
  publisher={Springer}
}

@article{cmipo2016scenario,
  title={The scenario model intercomparison project (ScenarioMIP) for CMIP6},
  author={O'Neill, Brian C and Tebaldi, Claudia and Van Vuuren, Detlef P and Eyring, Veronika and Friedlingstein, Pierre and Hurtt, George and Knutti, Reto and Kriegler, Elmar and Lamarque, Jean-Francois and Lowe, Jason and others},
  journal={Geoscientific Model Development},
  volume={9},
  number={9},
  pages={3461--3482},
  year={2016},
  publisher={Copernicus GmbH}
}

@article{stern2022time,
  title={A time for action on climate change and a time for change in economics},
  author={Stern, Nicholas},
  journal={The Economic Journal},
  volume={132},
  number={644},
  pages={1259--1289},
  year={2022},
  publisher={Oxford University Press}
}

@article{fussellogical,
  title={Logical and empirical flaws in applications of simple climate-economy models},
  author={F{\"u}ssel, HM},
  year={2006},
journal={Kluwer Academic Publishers},
  publisher={Kluwer Academic Publishers}
}

@article{mastrandrea2001integrated,
  title={Integrated assessment of abrupt climatic changes},
  author={Mastrandrea, Michael D and Schneider, Stephen H},
  journal={Climate Policy},
  volume={1},
  number={4},
  pages={433--449},
  year={2001},
  publisher={Taylor \& Francis}
}

\end{document}